\documentclass[10pt,twocolumn,letterpaper]{article}

\usepackage{cvpr}              

\usepackage[accsupp]{axessibility} 

\usepackage[dvipsnames]{xcolor}

\newcommand{\cparagraph}[1]{{\vspace{+1mm}\noindent\textbf{#1}\quad}}

\definecolor{cvprblue}{rgb}{0.21,0.49,0.74}
\usepackage[pagebackref,breaklinks,colorlinks,citecolor=cvprblue]{hyperref}
\usepackage{float}
\usepackage{marvosym}

\def\paperID{2291} 
\def\confName{CVPR}
\def\confYear{2024}

\title{EpiDiff: Enhancing Multi-View Synthesis via Localized Epipolar-Constrained Diffusion}

\author{
    {Zehuan Huang\textsuperscript{1}\footnotemark[1] \quad
    Hao Wen\textsuperscript{1}\footnotemark[1] \quad
    Junting Dong\textsuperscript{2}\footnotemark[1] \quad
    Yaohui Wang\textsuperscript{2} \quad
    Yangguang Li\textsuperscript{3} \quad
    Xinyuan Chen\textsuperscript{2}}\\
    \vspace{0.5em}{Yan-Pei Cao\textsuperscript{3} \quad
    Ding Liang\textsuperscript{3} \quad
    Yu Qiao\textsuperscript{2} \quad
    Bo Dai\textsuperscript{2}\footnotemark[2] \quad
    Lu Sheng\textsuperscript{1}\footnotemark[2]} \\
    {\textsuperscript{1}Beihang University \quad
    \textsuperscript{2}Shanghai AI Laboratory \quad
    \textsuperscript{3}VAST}
}

\begin{document}

\twocolumn[\maketitle\vspace{-2.5em}\begin{center}
\centering
\includegraphics[width=\textwidth]{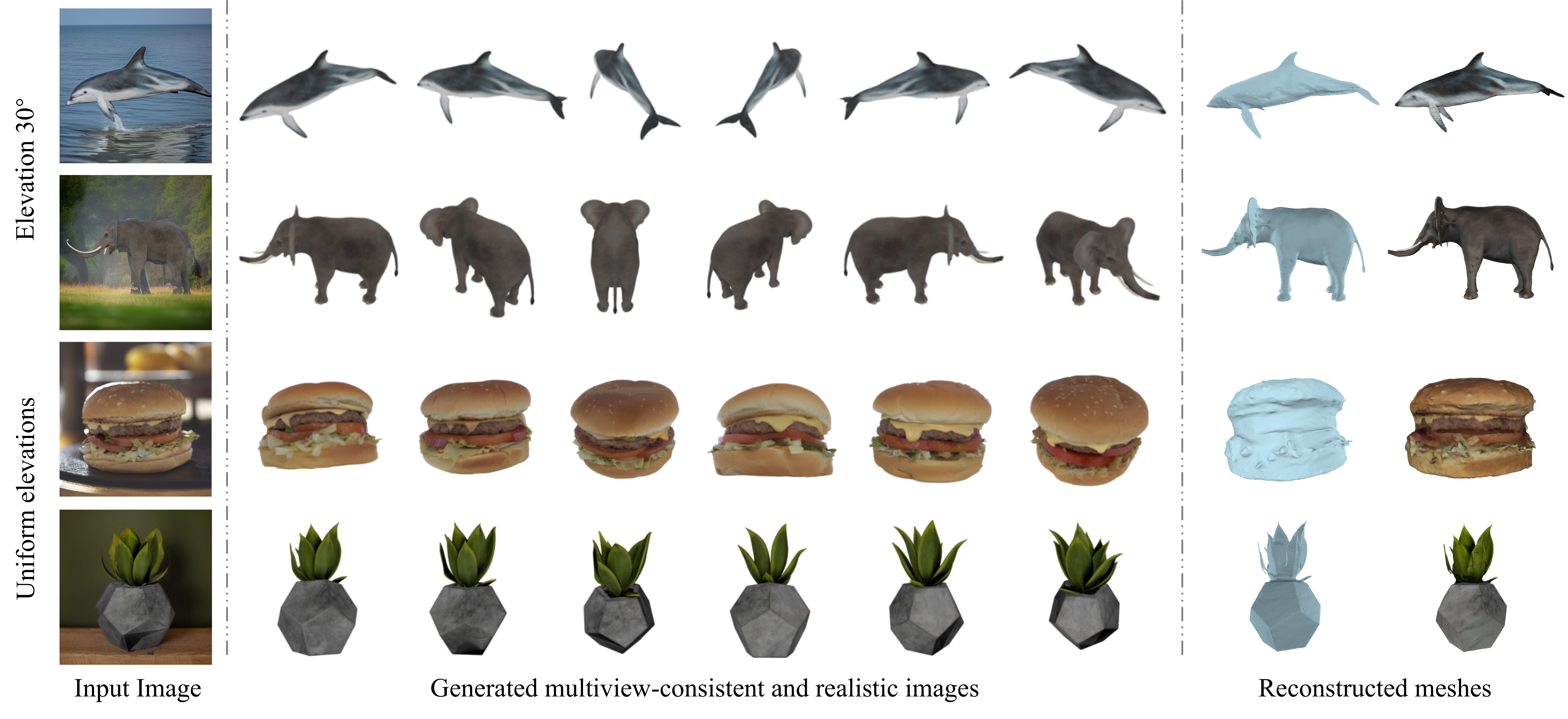}
\captionof{figure}{\textbf{EpiDiff} is able to efficiently generate multi-view consistent and high-quality images from a single view input. Instead of limited to fixed viewpoints, EpiDiff can generate relatively arbitrary multi-views. EpiDiff is lightweight and only takes 12 seconds to generate 16 multi-view images. The generated multiview images can be used to recover 3D shapes by neural reconstruction methods~\cite{wang2021neus,instant-nsr-pl}. }
\label{fig:header}
\end{center}\bigbreak]

\renewcommand{\thefootnote}{\fnsymbol{footnote}}
\footnotetext[1]{Equal Contribution}
\footnotetext[2]{Corresponding author}

\begin{abstract}
Generating multiview images from a single view facilitates the rapid generation of a 3D mesh conditioned on a single image. Recent methods~\cite{syncdreamer} that introduce 3D global representation into diffusion models have shown the potential to generate consistent multiviews, but they have reduced generation speed and face challenges in maintaining generalizability and quality.
To address this issue, we propose EpiDiff, a localized interactive multiview diffusion model. At the core of the proposed approach is to insert a lightweight epipolar attention block into the frozen diffusion model, leveraging epipolar constraints to enable cross-view interaction among feature maps of neighboring views. The newly initialized 3D modeling module preserves the original feature distribution of the diffusion model, exhibiting compatibility with a variety of base diffusion models. Experiments show that EpiDiff generates 16 multiview images in just 12 seconds, and it surpasses previous methods in quality evaluation metrics, including PSNR, SSIM and LPIPS. Additionally, EpiDiff can generate a more diverse distribution of views, improving the reconstruction quality from generated multiviews. Please see the project page at \href{https://huanngzh.github.io/EpiDiff/}{huanngzh.github.io/EpiDiff/}.
\end{abstract}

\section{Introduction}
\label{sec:intro}

3D reconstruction from a single image has a wide range of applications in augmented/virtual reality, robotic simulation, and gaming. However, this task presents significant challenges due to the complexity of accurately interpreting spatial structure and details from a 2D perspective observation. It requires extensive knowledge of the 3D world.

Recently, with the introduction of diffusion models~\cite{ho2020denoising,ldm_highresolution}, the field of 3D generation has experienced rapid and vigorous development. A series of research~\cite{dpm_pcd,point-e,lion,get3d,mesh-diffusion,renderdiffusion,single-stage-diff-nerf,sdfusion,controllable3d,3dgen,neuralfield-ldm,diffrf,autodecoding,rodin,shap-e,push3dshape,difftf} attempts to train 3D generative diffusion models from scratch on 3D assets. However, due to the limited size of publicly available 3D datasets, these methods demonstrate poor generalizability. Therefore, another group of works~\cite{dreamfield,dreamfusion,sjc,fantasia3d,prolificdreamer,realfusion,magic123,dreambooth3d,makeit3d,yu2023csd,sun2023dreamcraft3d} propose to distill prior knowledge of 2D diffusion models to create 3D models from text or images via Score Distillation Sampling (SDS)~\cite{dreamfusion}. Although these methods have achieved impressive results, they require expensive per-scene optimization that usually takes tens of minutes or even hours.

More recently, several works~\cite{zero123, syncdreamer} have emerged that finetune pretrained 2D image diffusion models~\cite{ldm_highresolution} on 3D datasets to generate novel view images, which can be utilized for 3D shape and appearance recovery with reconstruction methods~\cite{mildenhall2020nerf,wang2021neus}. This feed-forward pipeline not only retains generalization capacity of image diffusion models, but also allows for faster synthesis of multiview images. Zero123~\cite{zero123} proposes a view-conditioned diffusion model by embedding view features and camera information. However, the multiview consistency of generated images is limited, preventing high-quality 3D shape reconstruction. To improve the multiview consistency, SyncDreamer~\cite{syncdreamer} introduces a 3D feature volume constructed with global multiview images into the Zero123 backbone. However, the heavy global feature volume modeling not only slows down the generation process but also easily overfits to specific viewpoints and tends to produce low-quality images with distorted appearances and blurriness (see \cref{fig:mvs-elevation-30}).

In this paper, we present EpiDiff, a novel localized interactive multiview diffusion model. The key idea is to utilize the feature maps of neighboring views that suffer fewer occlusions for the generation of target view images, which improves the generalization capability and efficiency. To fuse the neighboring feature maps, during the generation of $N$ target views, we insert a lightweight epipolar attention block into the UNet, which leverages epipolar constraints to enable interaction among feature maps of $F$ neighboring views within a smaller range. The 3D modeling module preserves the original feature distribution of the diffusion model, exhibiting compatibility with various base diffusion models. Compared to global modeling methods, our localized modeling approach not only speeds up the generation process, but also facilitating the generation of more diversely distributed views, contributing to improved reconstruction from generated multiviews.

We conduct extensive experiments on the Google Scanned Object dataset~\cite{gso} and various styles of 2D images. The experiments validate that, compared to baseline methods, EpiDiff can generate higher quality multiview images more rapidly. The contributions of EpiDiff are summarized as follows:

\begin{itemize}
\item EpiDiff employs a 3D modeling module into the frozen diffusion model. The module preserves the original feature distribution of the diffusion model, exhibiting compatibility with various diffusion models.
\item We propose an epipolar attention block to learn the inter-correlations among neighboring views based on epipolar geometry relationships. This localized interactive and lightweight module models consistency effectively.
\item Experiments show that EpiDiff generates 16 multiviews in just 12 seconds, and it surpasses previous methods in metrics including PSNR, SSIM, and LPIPS. Additionally, it can generate more freely distributed views, improving the reconstruction quality from generated multiviews.
\end{itemize}

\section{Related work}
\label{sec:related}

\cparagraph{3D diffusion models} Recently, the field of 3D generation has witnessed rapid and significant advancements with the introduction of diffusion models. Motivated by the impressive generative performance of diffusion models trained on extensive image datasets, several studies have explored training 3D generative diffusion models from scratch on 3D assets. These 3D diffusion models directly generate point clouds~\cite{dpm_pcd,point-e,lion}, meshes~\cite{get3d,mesh-diffusion}, and neural radiance fields~\cite{renderdiffusion,single-stage-diff-nerf,sdfusion,controllable3d,3dgen,neuralfield-ldm,diffrf,autodecoding,rodin,shap-e,push3dshape,difftf} through network inference. However, owing to the considerably smaller size of publicly accessible 3D datasets compared to image datasets, these methods exhibit limited generalizability. They often generate shapes restricted to specific categories and face challenges in adapting to more intricate input conditions.

\cparagraph{3D generation guided by 2D prior} Given the robust generative capabilities of 2D diffusion models~\cite{ldm_highresolution,dalle2,imagen}, some research efforts directly leverage pretrained image diffusion models. With DreamFusion~\cite{dreamfusion} and SJC~\cite{sjc} proposing to distill a pretrained 2D text-to-image generation model for generating 3D shapes, a series of subsequent studies~\cite{magic3d,fantasia3d,prolificdreamer,realfusion,magic123,dreambooth3d,makeit3d,yu2023csd,sun2023dreamcraft3d,guo2023threestudio,yu2023hifi,cao2023dreamavatar} adopt such a pre-scene optimization pipeline. These methods optimize a 3D representation (e.g., NeRF, mesh, or SDF) conditioned on text or images through Score Distillation Sampling (SDS)~\cite{dreamfusion}. However, the per-scene optimization approach has issues with efficiency. Generating a single scene may take tens of minutes or even hours.

\cparagraph{Using 2D diffusion for multiview synthesis}
2D diffusion models have introduced significant progress in generating multiview images from a single view. While some approaches~\cite{nvs-3ddiff,nerdi,nerfdiff,mvdiffusion,poseguideddiff,nvsdiff,3dawarediff,longtermdiff,gao2024genesistex} integrate 3D representations or utilize shared attention for multiview or texture synthesis, they lack design for 3D reconstruction. Zero123~\cite{zero123} introduces relative view condition to diffusion models, enabling novel view synthesis (NVS) from a single image. However, due to its limited consistency, it utilizes SDS loss~\cite{dreamfusion} to achieve 3D reconstruction.
One-2-3-45~\cite{one2345} leverages a generalizable neural reconstruction method~\cite{long2022sparseneus} for producing 3D shapes from multiview images generated by Zero123~\cite{zero123}, yet the resultant quality is suboptimal.
More recently, SyncDreamer~\cite{syncdreamer} has introduced a 3D global feature volume during multiview generation to ensure consistency. However, the heavy global 3D representation modeling not only slows down the generation but also tends to produce low-quality images. In contrast, our localized modeling method facilitates the generation of multiview-consistent and high-quality images.

\cparagraph{Generalizable radiance field} Since NeRF~\cite{mildenhall2020nerf} achieved advancements in fitting single scene for novel view synthesis, numerous studies have explored generalizable neural radiance field representation for various scenes. These methods take a few input view images and generate a 3D implicit field in a single feed-forward pass, eliminating the need for per-scene optimization.
Among the methods, some explicitly construct the 3D feature or cost volume~\cite{mvsnerf,johari2022geonerf,zhang2022nerfusion,long2022sparseneus} and utilize the voxel feature for decoding density and color, inspiring SyncDreamer\cite{syncdreamer} to model 3D global representation, while others~\cite{wang2021ibrnet,reizenstein2021common,henzler2021unsupervised,2021pixelnerf,yang2023contranerf,liu2021neural,kulhanek2022viewformer,trevithick2020grf,sajjadi2022object,sajjadi2022srt,suhail2022gpnr,cheng2023sparsegnv} directly aggregate 2D features using MLPs or transformers, guided by spatial projection relationships. These methods, either explicitly or implicitly modeling generalized scenes, provide valuable insights for consistency modeling in the feed-forward multiview generation process.

\section{Method}
\label{sec:method}

\begin{figure*}
    \centering
    \includegraphics[width=0.95\textwidth]{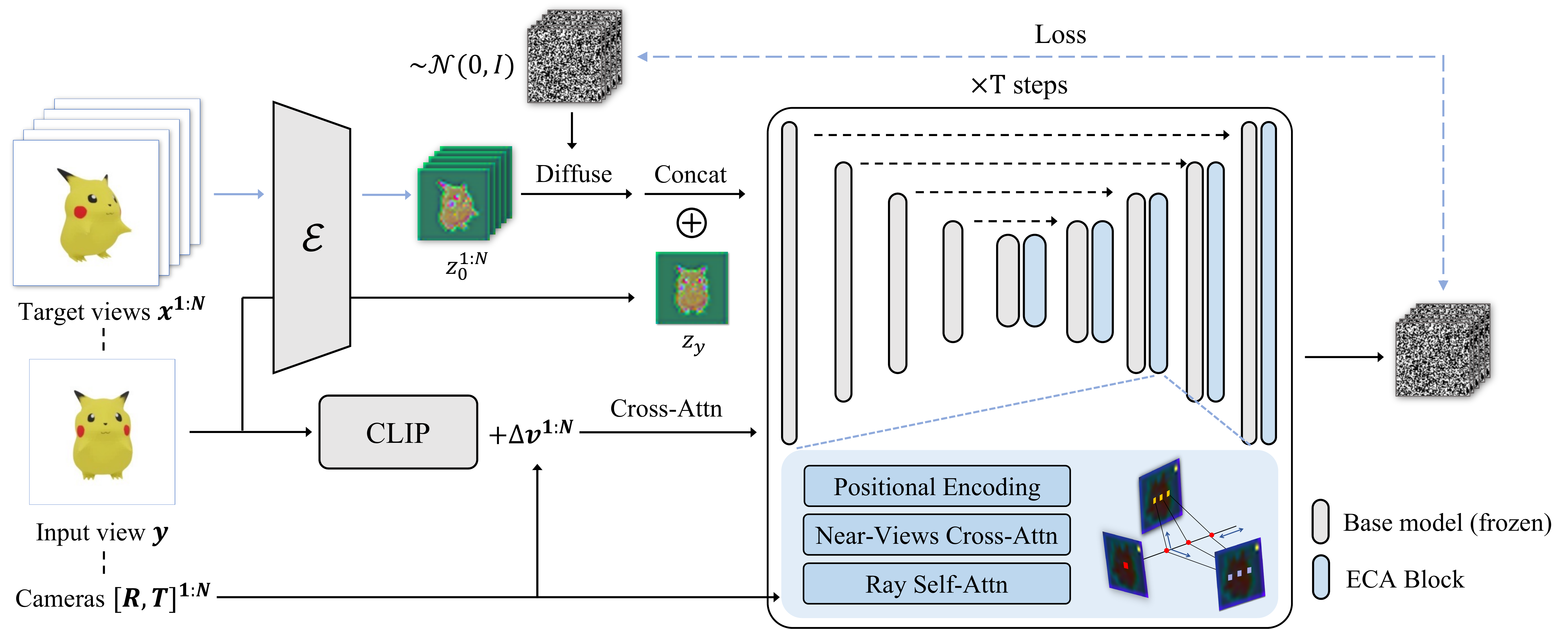}
    \caption{\textbf{Pipeline of EpiDiff.} Based on a base NVS model (e.g., Zero123~\cite{zero123}), our method designs a module for modeling 3D consistency, which is inserted into the mid-sample and up-sample stages of the UNet. We use attention networks to construct the module, aimed at learning generalized epipolar geometry, termed Epipolar-constrained Attention Block (ECA Block). During training, only parameters of the ECA Block are updated, thereby preserving the feature space of the base model and encouraging the module to extract 3D priors.}
    \label{fig:pipeline}
\end{figure*}

Given an input view $y$ of an object, our goal is to generate multiview images of the object. Let $R\in \mathbb{R}^{3\times 3}$ and $T\in \mathbb{R}^{3}$  represent the rotation and translation of the camera relative to the input view for the target viewpoints, respectively. We denote the target viewpoints for the $N$ views as $[R , T]^{1:N}$. Therefore, the aim is to learn a model $f$ that accepts the input view and synthesizes multiview images under specified camera transformations:
\begin{equation}
    \hat{x}^{1:N} = f(y, [R , T]^{1:N})
    \label{equ:goal}
\end{equation}
where $\hat{x}^{1:N}$ is denoted as the generated multiview images.


To enhance the consistency, quality and efficiency of the multiview generation, we introduce a localized interactive multiview diffusion scheme based on the novel view synthesis base model. We design a lightweight attention block in the original feature space of the diffusion model, with the aim of establishing nearby view associations to model 3D representation.
As illustrated in \cref{fig:pipeline},
the proposed approach involves an interaction process among the intermediate features of multiview images (refer to \cref{sec:mvdiffusion}), thereby extending single-view synthesis to multiview synthesis. Specifically, we design a 3D modeling module, named Epipolar-constrained Attention Block (ECA Block) based on the attention mechanism and insert it into the base diffusion model (refer to \cref{sec:eca_block}). The ECA Block facilitates information exchange among latent features of  neighboring views, leveraging epipolar constraints in stereo vision to enhance consistency and quality in multiview images. The ECA Block, with its lightweight and adaptable characteristics, is suitable for various base models.

\subsection{Multiview diffusion model}
\label{sec:mvdiffusion}

\cparagraph{Base diffusion model} The proposed multiview diffusion model is rooted in image diffusion models, typically constructed on the Latent Diffusion Models (LDM)~\cite{ldm_highresolution}. The LDM operates the denoising process in the latent space of an autoencoder, namely $\mathcal{E}(\cdot)$ and $\mathcal{D}(\cdot)$. During the training phase, an input image $x_{0}$ is initially mapped to the latent space using a frozen encoder, yielding $z_{0}=\mathcal{E}(x_{0})$, then perturbed by a pre-defined Markov process:
\begin{equation}
    q(z_{t}|z_{t-1})=\mathcal{N}(z_{t}; \sqrt{1-\beta_{t}}z_{t-1}, \beta_{t}I)
    \label{eq:img_diff}
\end{equation}
for $t=1,\cdots,T$, with $T$ represents the number of steps in the forward diffusion process. The sequence of hyperparameters $\beta_{t}$ determines the noise strength at each step. The denoising UNet $\epsilon_{\theta}$, featuring four downsample blocks, one middle block and four upsample blocks to attain four resolution levels, is trained to approximate the reverse process $q(z_{t-1}|z_{t})$. The vanilla training objective is expressed as:
\begin{equation}
    \mathcal{L}=\mathbb{E}_{\mathcal{E}(x_{0}),\epsilon\sim\mathcal{N}(0,I),t}[\lVert \epsilon-\epsilon_{\theta}(z_{t},t) \rVert_{2}^{2}]
    \label{eq:img_diff_loss}
\end{equation}

Zero123~\cite{zero123}, a base model for novel view synthesis (NVS), employs an input view $y$ and camera parameters $[R,T]$ as conditions to extend from image to 3D domain. This model, built on a pretrained image diffusion model and augmented with camera control, aims to optimize the following objective:
\begin{equation}
    \mathcal{L}=\mathbb{E}_{\mathcal{E}(x_{0}),\epsilon\sim\mathcal{N}(0,I),t}[\lVert \epsilon-\epsilon_{\theta}(z_{t},t,c(y,R,T)) \rVert_{2}^{2}]
    \label{eq:zero123_loss}
\end{equation}
where $c(x,R,T)$ is the embedding of the input view and relative camera extrinsics.

\cparagraph{Interactive multiview diffusion} Drawing upon the foundational image or NVS diffusion models, our multiview diffusion model promotes spatial interaction among images at the latent feature level. It incorporates 3D geometric significance during the step-by-step denoising process, thus generating 3D-consistent multiview images.

Given a dataset of multiview images and their associated camera extrinsics $\{(x^{1:N}, [R,T]^{1:N})\}$, we finetune a foundational diffusion model to learn 3D consistency. $x^{1}$ is selected as the input view image $y$ in the training stage. Multiview images $x^{1:N}$ are mapped to latent codes $z_{0}^{1:N}$ through the VAE encoder $\mathcal{E}$.

We introduce a generalized 3D consistency module into the UNet of the base model. This module, inserted into the middle and each upsample block of the UNet, allows multiview image feature maps to interact in 3D space, creating consistent interconnections. The integration of the module transforms the generation of multiple images from independent processes into a synchronous interactive process. The final denoising network $\epsilon_{\theta}$ is trained on a dataset comprising multiview images and camera extrinsics. The optimization objective is defined as:
\begin{equation}
    \mathcal{L}=\mathbb{E}_{\mathcal{E}(x_{0}^{1:N}),\epsilon\sim\mathcal{N}(0,I),t}[\lVert \epsilon-\epsilon_{\theta}(z_{t}^{1:N},t,y,[R,T]^{1:N}) \rVert_{2}^{2}]
    \label{eq:mvs_diff_loss}
\end{equation}

By freezing the base model and exclusively training the 3D modeling module, we maintain the original generative capabilities of the base model. This add-on module, modeling multiview associations, enhances the base diffusion models' ability to generate multiview-consistent images. Our design of the multiview diffusion framework is adaptable to various base diffusion models.

\subsection{Epipolar-constrained attention}
\label{sec:eca_block}
We introduce a module that uses a series of attention networks for cross-view interaction, in accordance with the principles of 3D projection. This module facilitates the learning of a generalizable epipolar geometry, which we refer to as the Epipolar-constrained Attention Block (ECA Block). The ECA Block identifies correlations between latent features of the current view and corresponding features along the epipolar line in other views. 

\begin{figure}[!t]
\centering
\includegraphics[width=\linewidth]{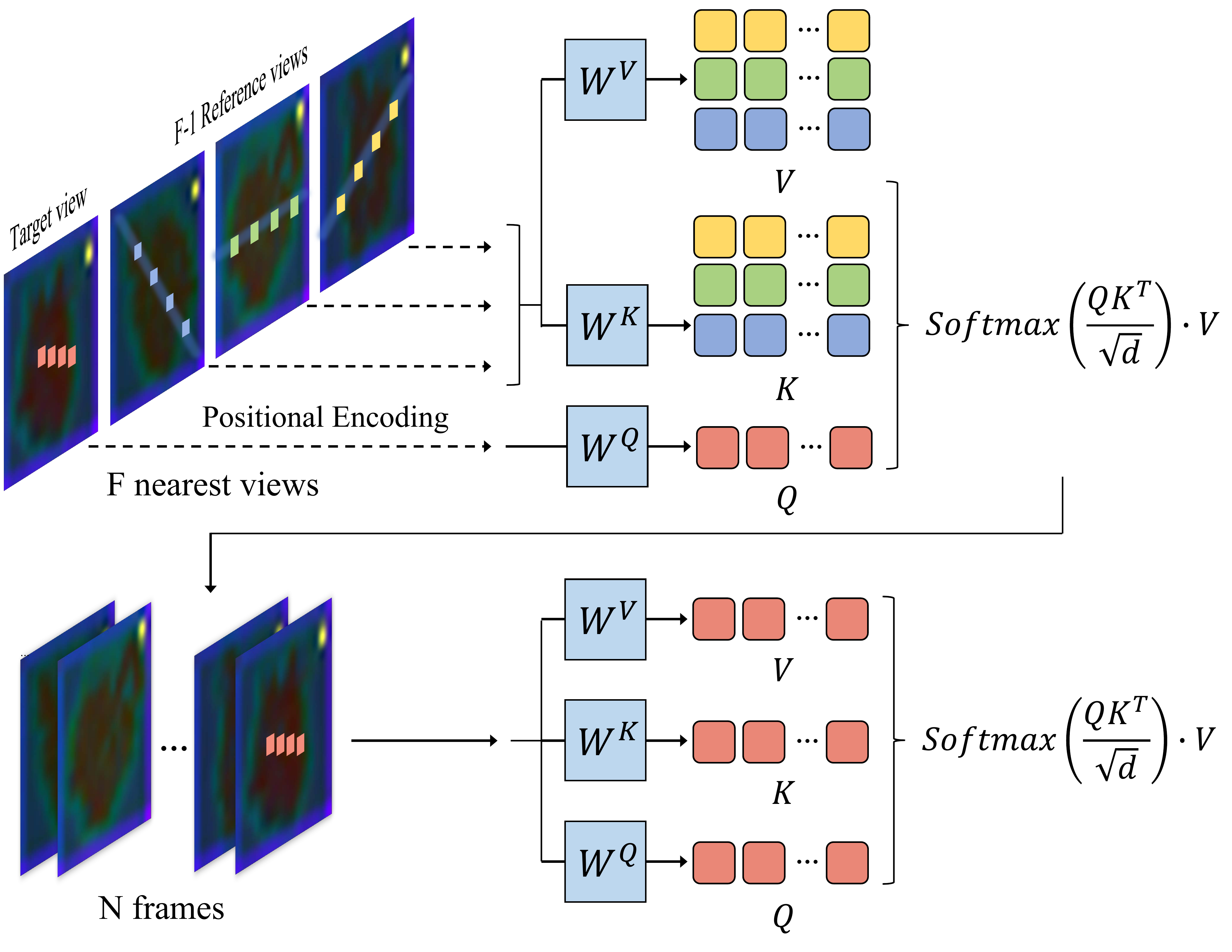}
\caption{\textbf{Illustration of our Epipolar-constrained Attention Block (ECA Block).} Latent features of multiview images are associated in 3D space through two key attention blocks. The Near-Views Cross-Attn initially aggregates features from nearby views onto the target view's ray points, guided by epipolar geometry. Subsequently, Ray Self-Attn models depth information to fuse ray features into 2D feature maps. The ECA Block facilitates multiview interaction by effectively harnessing spatial geometry.}
\label{fig:ecablock}
\end{figure}

As illustrated in \cref{fig:ecablock}, the ECA Block comprises two primary attention blocks, supplemented with linear layers: (i) the Near-Views Cross-Attention module, tasked with aggregating epipolar line features from neighboring views for current view ray features, and (ii) the Ray Self-Attention module, designed to model the depth weights of the sampled rays, fuse spatial points into pixel features.

\cparagraph{Near-Views Cross-Attention} Given a target view, our method relies on the features of the $F$ closest views among the total $N$ views to produce the output feature map. We start by emitting rays from patches on the target view, and sample $S$ points uniformly along each ray. We then project them onto $F-1$ neighboring views. For each point $s$, we get the feature $v^{s}$ by positional encoding and projected feature $\{f^{s}_{i} \mid 1\leq i\leq F-1\}$ on reference views. We enable cross-view interaction by a cross-attention layer. Specifically, using point feature $v^{s}$ as query, we implement $\mathrm{Attention}(Q,K,V)=\mathrm{Softmax}(\frac{QK^T}{\sqrt{d}})\cdot V$ with
\begin{equation}
    Q=W^{Q}v^{s}, K=W^{K}[f^{s}_{1},\cdot\cdot,f^{s}_{F-1}], V=W^{V}[f^{s}_{1},\cdot\cdot,f^{s}_{F-1}],
    \label{eq:near-view-cross-attn}
\end{equation}
where $[\cdot]$ denotes concatenation operation, and $W^{Q}$, $W^{K}$, $W^{V}$ are learnable matrices that project the inputs to query, key, and value, respectively.


During the synthesis of the target view feature map, features from nearby views with a 3D geometric relationship to the target view are integrated, as described in \cref{eq:near-view-cross-attn}. This approach effectively captures inter-view feature correlations, thereby enhancing multiview consistency efficiently.


\cparagraph{Ray Self-Attention} To model spatial points' contributions to pixel-level features, the ray aggregation module is implemented using self-attention. Upon acquiring the feature of each sampled point $\widetilde{v}^{s}(1\leq s\leq S)$, we apply self-attention along the ray with dimension of $S$, deriving all of query, key and value features from $[\widetilde{v}^{1},\cdots,\widetilde{v}^{S}]$. This step is crucial for revealing depth relationships among  sampled points along rays and capturing their spatial correlations. Finally, we perform a weighted summation to fuse 3D features into 2D feature of target patch $\widetilde{f}$.

Our method aligns features precisely within four resolution levels of the UNet latent space by leveraging epipolar projection, which preserves the feature integrity of the base model. The subdivision of the 3D attention process into two modules further allows for efficient multiview consistency modeling.

\cparagraph{Positional Encoding} To enhance the module's understanding of perspective relationships, we have adopted a positional encoding technique inspired by Light Field Networks (LFN)~\cite{LFN}. This approach is distinct in that each pixel is associated with a ray, utilizing Plücker coordinates for representation, denoted as $r = (o\times d, d)$, where $o$ signifies the ray's origin, and $d$ its direction. Unlike traditional position encoding methods where all pixels of an image share the same positional encoding, the LFN~\cite{LFN} method employs MVP transformation to back-project each pixel of the image into the world coordinate system, precisely locating the corresponding ray for each pixel. It introduces strict geometric constraints and sufficiently incorporates camera parameters, effectively applying encoding at the pixel level. By integrating these encoded coordinates obtained by LFN with latent feature maps via a harmonic transformation, our method significantly enriches the spatial information within the feature representation.

Furthermore, to address the potential biases caused by using absolute positions, we also adopt the strategy from GPNR~\cite{suhail2022gpnr}, establishing a ray-relative positioning system through its corresponding camera extrinsics $[R,T]$. We construct canonicalizing transformation $\mathcal{T}$ using the Y-axis of $R$ ($y$) and the ray direction ($v$) as the Z-axis. Let $v' = \frac{v}{\|v\|}$ and $y'=y-(y\cdot v')v'$ , the transformation $\mathcal{T}$ is formed as:
\begin{align}
R_c &= \left[ \frac{y'}{\|y'\|} \times v' \quad \frac{y'}{\|y'\|} \quad v' \right] \\
\mathcal{T} &= \left[ R_c^{\top} \ | \ -R_c^{\top} T \right]
\end{align}
By applying $\mathcal{T}$ to each ray, we adjust them to 
originate from \( (0,0,0) \) and align along the direction \( (0,0,1) \). Finally, we transform the rays from all neighboring viewpoints into the relative coordinate system. It ensures that the rays are consistently represented relative to a common reference, mitigating biases from absolute positioning.
\section{Experiments}
\label{sec:experiments}

\subsection{Experiment settings}
\cparagraph{Dataset} Our model is trained on the LVIS subset of the Objaverse~\cite{objaverse}, comprising over 40,000 3D models. For each object, we setup a camera layout to uniformly sample 96 views with a size of 256$\times$256. The layout comprises six concentric circles, each as a layer, with elevation angles set at  $\{-10^{\circ}, 0^{\circ}, 10^{\circ}, 20^{\circ}, 30^{\circ}, 40^{\circ}\}$. Each circle contains 16 cameras spaced evenly from $0^{\circ}$ to $360^{\circ}$ in azimuth angle. During training, 16 views from each model are randomly sampled.

We evaluate EpiDiff using the Google Scanned Object dataset~\cite{gso}, which consists of over one thousand scanned models. To assess multiview synthesis, we employ two rendering settings. Firstly, aligning with SyncDreamer's training setting~\cite{syncdreamer}, we fix elevation at $30^{\circ}$ and sample 16 cameras with azimuths evenly spread from $0^{\circ}$ to $360^{\circ}$. Secondly, we render 16 views with azimuths also uniformly distributed in $[0^{\circ}, 360^{\circ}]$, but with varying elevations from $-10^{\circ}$ to $40^{\circ}$. This tests the model's ability to generate multiview images across a wider range of viewpoints.

\cparagraph{Implementation details} We use Zero123~\cite{zero123} as our backbone. EpiDiff is trained for 30,000 steps (around 2 days) using 8 80G A800 GPUs, with a learning rate of 1e-5 and a total batch size of 512. In each iteration, we generate 16 multiview images, i.e., $N=16$. Within the ECA Block, the number of sampled rays is aligned with the resolution of feature maps and the number of sampled points per ray $S$ is set to 16. For the Near-Views Cross Attention module, the number of nearby views $F$, is set to 4. For reconstruction, the foreground masks of the generated images are initially predicted using CarveKit. Then, we use the instant-NGP based SDF reconstruction method~\cite{instant-nsr-pl} to reconstruct the shape. Each object undergoes training for 3k steps, which costs about 2 minutes.

\cparagraph{Baselines} We adopt Zero123~\cite{zero123}, One-2-3-45~\cite{one2345}, Point-E~\cite{point-e}, Shap-E~\cite{shap-e}, SyncDreamer~\cite{syncdreamer} as baseline methods. We adopt the official implementation of Zero123 XL\cite{zero123}. One-2-3-45~\cite{one2345} uses Zero123 to synthesize multiview images and regresses SDFs from them. Point-E~\cite{point-e} and Shap-E~\cite{shap-e} are 3D generative models trained on a large internal OpenAI 3D dataset. We adopt the best released version of Point-E, \textit{base1B}, which is a 1 billion parameter image to point cloud diffusion model. We use it to generate point clouds, then convert the generated point clouds to SDFs using the official models. For Shap-E~\cite{shap-e}, We use its STF rendering mode to predict SDF values and texture colors, enabling direct generation of colored vertex meshes. SyncDreamer~\cite{syncdreamer} builds a multiview shared 3D feature volume based on Zero123, aiming to generate multiview consistent images from a single input image. We utilize the official source code in our elevaluation.

\cparagraph{Metrics} We primarily focus on two tasks: multiview synthesis and single-view 3D reconstruction. For multiview synthesis, we employ the commonly used evaluation metrics, i.e., PSNR, SSIM~\cite{ssim}, and LPIPS~\cite{lpips} to assess our method's performance. In single-view 3D reconstruction task, we measure the quality of reconstructed shapes using Chamfer Distance and Volumetric Intersection over Union (Volume IoU) compared to the ground truth. For comparing reconstruction performance between Zero123~\cite{zero123}, SyncDreamer~\cite{syncdreamer}, and our method EpiDiff, we use multiview images generated under the fixed elevation $30^{\circ}$ setting.

\subsection{Consistent multiview synthesis}

\begin{figure*}
\centering
\includegraphics[width=0.95\textwidth]{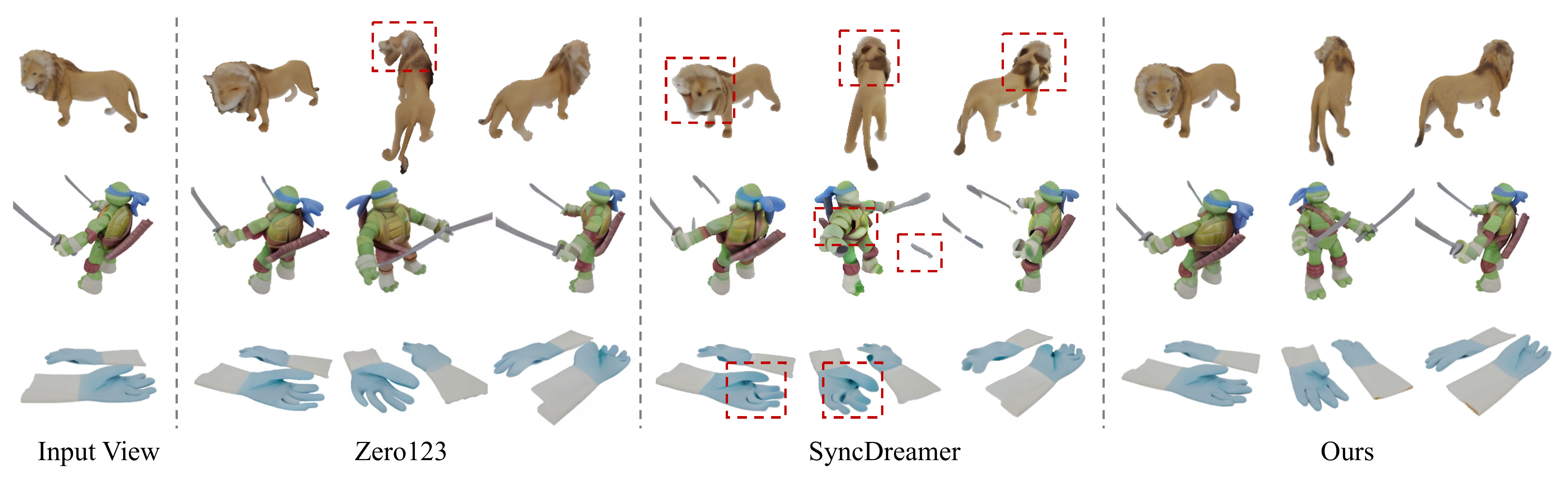}
\caption{Qualitative comparison with Zero123~\cite{zero123} and SyncDreamer~\cite{syncdreamer} under the elevation $30^{\circ}$ setting.}
\label{fig:mvs-elevation-30}
\end{figure*}

\begin{figure*}
\centering
\includegraphics[width=0.95\textwidth]{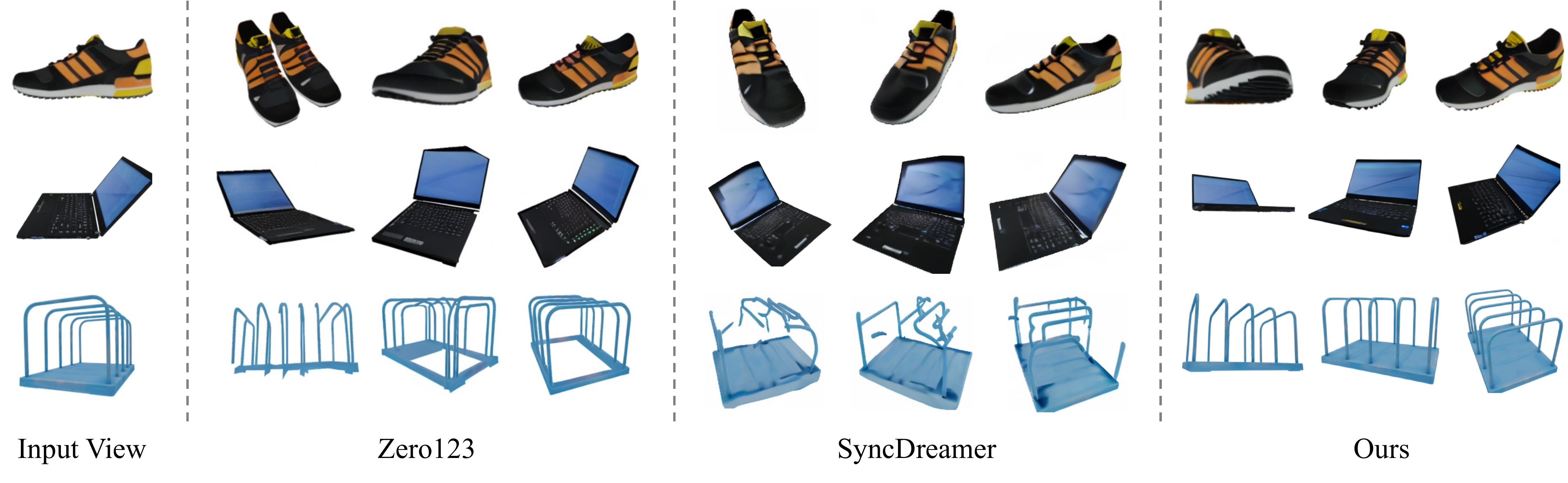}
\caption{Qualitative comparison with Zero123~\cite{zero123} and SyncDreamer~\cite{syncdreamer} under the uniform elevation setting. We present the generated results of each method when the elevation is $-10^{\circ}$, $10^{\circ}$ and $30^{\circ}$.}
\label{fig:mvs-elevation-uniform}
\end{figure*}

\begin{figure*}
\centering
\includegraphics[width=0.95\textwidth]{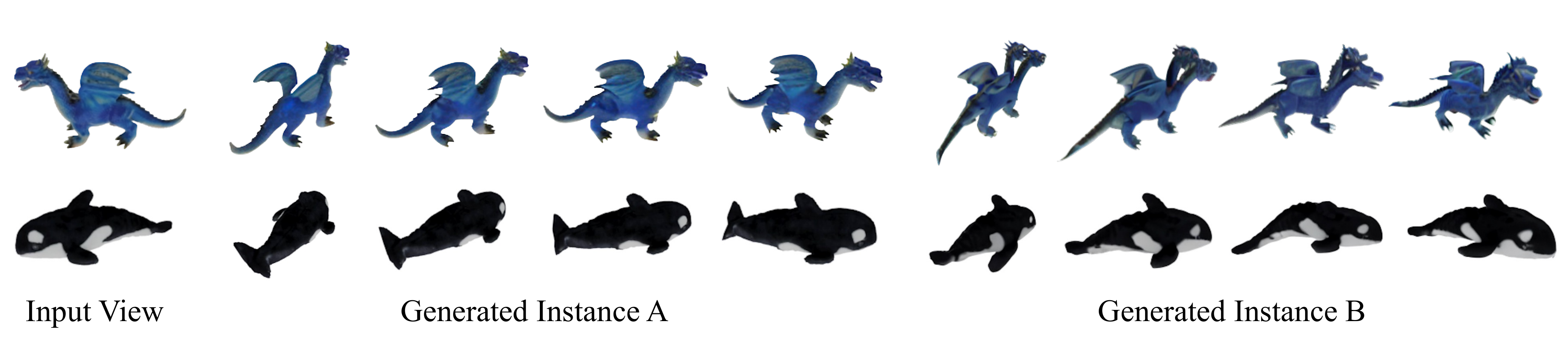}
\caption{Different plausible instances generated by EpiDiff from the same input image.}
\label{fig:diversity}
\end{figure*}

\begin{table}\small
  \centering
  \begin{tabular}{@{}lcccc@{}}
    \toprule
    Method & PSNR$\uparrow$ & SSIM$\uparrow$ & LPIPS$\downarrow$ & Runtime$\downarrow$ \\
    \midrule
    Zero123~\cite{zero123} & 17.79 & 0.796 & 0.201 & 7s \\
    SyncDreamer~\cite{syncdreamer} & 20.11 & 0.829 & 0.159 & 60s \\
   Ours & \textbf{20.49} & \textbf{0.855} & \textbf{0.128} & 12s \\
    \bottomrule
  \end{tabular}
  \caption{Quantitative comparison in multiview synthesis under elevation $30^{\circ}$ setting. We report PSNR, SSIM~\cite{ssim}, LPIPS~\cite{lpips}, Runtime on the GSO~\cite{gso} dataset.}
  \label{tab:mvs-elevation-30}
\end{table}

\begin{table}\small
    \centering
    \begin{tabular}{@{}lcccc@{}}
   \toprule
   Method & PSNR$\uparrow$ & SSIM$\uparrow$ & LPIPS$\downarrow$ & Runtime$\downarrow$ \\
   \midrule
   Zero123~\cite{zero123} & 15.91 & 0.772 & 0.231 & 7s \\
   SyncDreamer~\cite{syncdreamer} & 15.90 & 0.773 & 0.246 & 60s \\
   Ours & \textbf{18.83} & \textbf{0.821} & \textbf{0.163} & 12s \\
   \bottomrule
\end{tabular}
\caption{Quantitative comparison in multiview synthesis under uniform elevation setting. We report PSNR, SSIM~\cite{ssim}, LPIPS~\cite{lpips}, Runtime on the GSO~\cite{gso} dataset.}
\label{tab:mvs-elevation-uniform}
\end{table}

For this task, we use two viewpoint sampling settings. Under the setting with a fixed elevation of $30^{\circ}$, the quantitative results are shown in \cref{tab:mvs-elevation-30} and the qualitative results are shown in \cref{fig:mvs-elevation-30}. Zero123~\cite{zero123} generates visually plausible images, but the generated images are inconsistent. SyncDreamer~\cite{syncdreamer} generates more consistent multiview images, but the image quality is relatively poor, exhibiting blurry and unrealistic appearances. Our method efficiently generates multiview images that not only maintain semantic consistency with the input image, but also demonstrate higher performance in terms of structure and realism.

Under the uniform elevation setting, the quantitative results are shown in \cref{tab:mvs-elevation-uniform}, and the qualitative results are shown in \cref{fig:mvs-elevation-uniform}. The uniform view setting offers a more flexible viewpoint distribution. SyncDreamer~\cite{syncdreamer} lacks precise cross-view geometric constraints, leading to overfitting training viewpoints and a bias towards top-down perspectives in multiview generation, as shown in \cref{fig:mvs-elevation-uniform}. Our method can still generate relatively realistic and multiview-consistent images, benefiting from (a) the introduction of localized epipolar geometric constraints along with effective camera position encoding, and (b) the random viewpoint sampling training method. Meanwhile, for the same input image, Our method can generate different plausible instances using different random seeds as shown in \cref{fig:diversity}.

\subsection{Single view reconstruction}

\begin{table}\small
    \centering
    \begin{tabular}{@{}lcc@{}}
   \toprule
   Method & Chamfer Dist.$\downarrow$ & Volume IoU$\uparrow$ \\
   \midrule
   One-2-3-45~\cite{one2345} & 0.0768 & 0.2936 \\
   Point-E~\cite{point-e} & 0.0570 & 0.2027 \\
   Shap-E~\cite{shap-e} & 0.0689 & 0.2473 \\
   Zero123~\cite{zero123} & 0.0543 & 0.3358 \\
   SyncDreamer~\cite{syncdreamer} & 0.0496 & 0.4149 \\
   Ours & \textbf{0.0429} & \textbf{0.4518} \\
   \bottomrule
\end{tabular}
\caption{Quantitative comparison of surface reconstruction. We report Chamfer Distance and Volume IoU on the GSO dataset~\cite{gso}.}
\label{tab:recon-elevation-30}
\end{table}

\begin{figure*}
\centering
\includegraphics[width=0.95\textwidth]{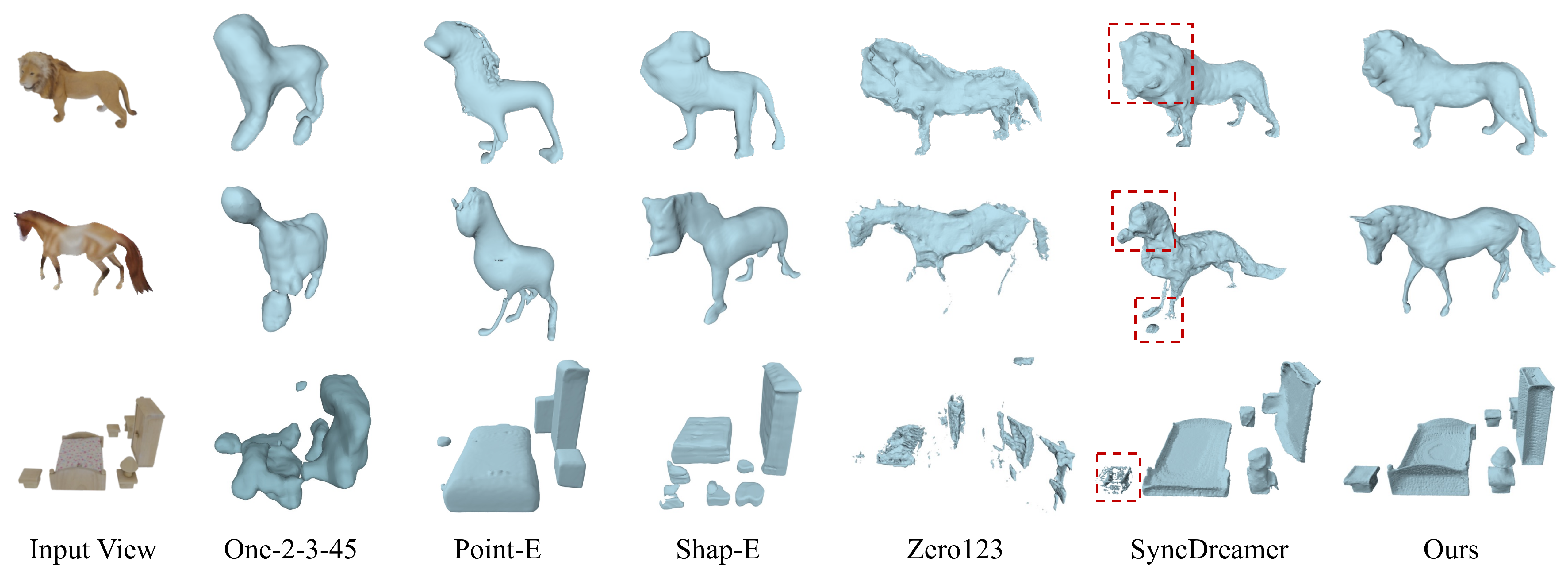}
\caption{Qualitative comparison of surface reconstruction from single image. For Zero123~\cite{zero123}, SyncDreamer~\cite{syncdreamer} and EpiDiff, we use multiview images generated under elevation $30^{\circ}$ setting.}
\label{fig:recon-elevation-30}
\end{figure*}

We present the quantitative results in \cref{tab:recon-elevation-30} and the qualitative results in \cref{fig:recon-elevation-30}. One-2-3-45~\cite{one2345} generates smooth meshes using a generalizable neural reconstruction approach~\cite{long2022sparseneus}, but lacks detail. Point-E~\cite{point-e} and Shap-E~\cite{shap-e} tend to generate incomplete shapes and exhibit differences from the objects shown in the input images. Zero123~\cite{zero123}, due to its poor multiview consistency, results in missing or rough geometry in the reconstruction. SyncDreamer~\cite{syncdreamer} generates a roughly consistent geometric framework but struggles to handle details. In comparison, our method achieves the best reconstruction quality in terms of smooth surfaces and detailed geometric structures.

\begin{figure}[ht]
    \centering
   \includegraphics[width=0.9\linewidth]{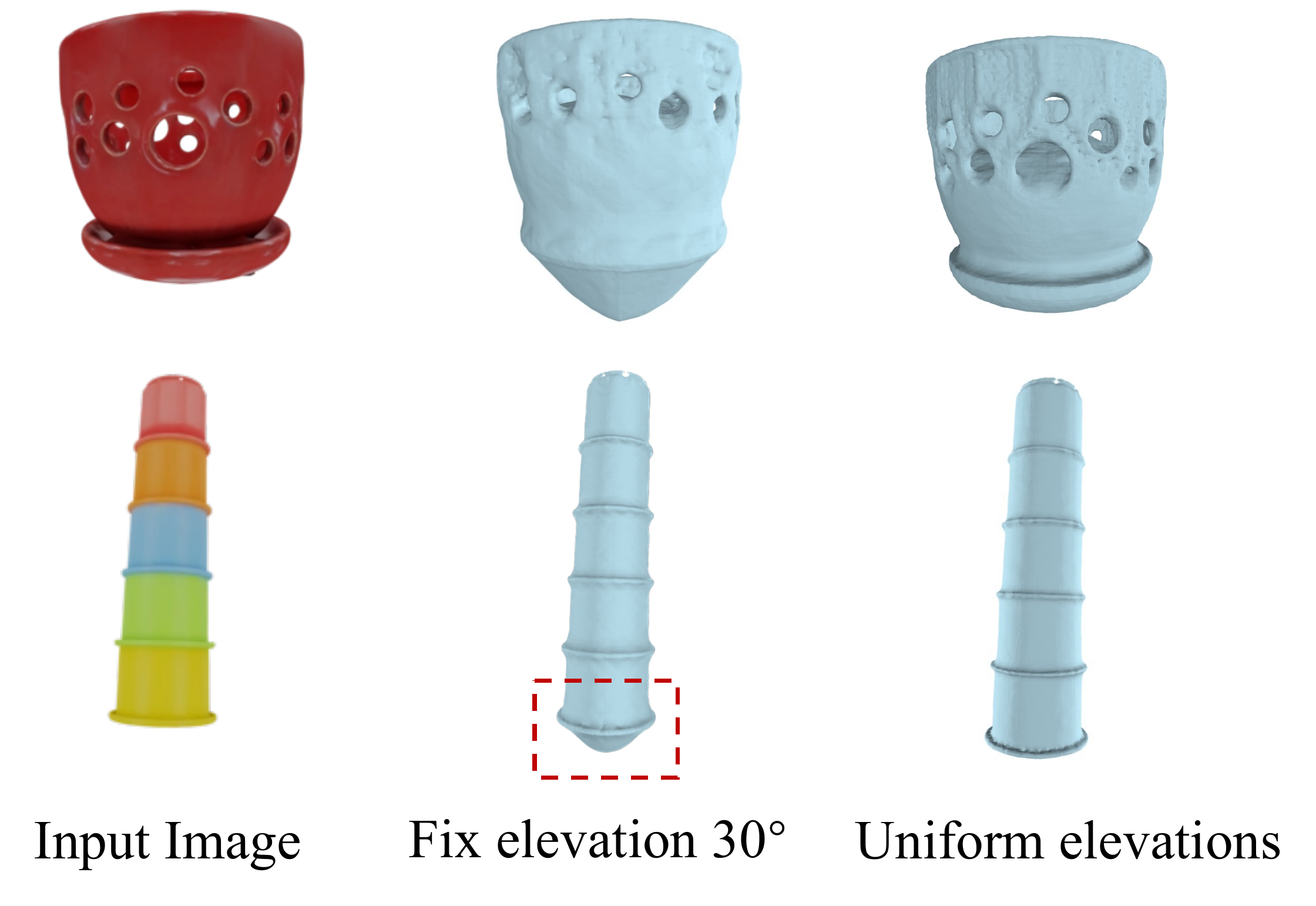}
   \caption{The common failure mode when reconstructing using multiviews at a fixed elevation of $30^{\circ}$. When the bottom of the object is round-shaped, the reconstructed bottom appears protruded because the camera’s visible frustum can not cover the bottom. Using a uniform distribution of elevations can solve it.}
   \label{fig:recon-comp-ele}
\end{figure}

We also observe, as shown in \cref{fig:recon-comp-ele}, that reconstructing using 16 generated views with a fixed elevation leads to unrealistic shapes at the bottom of objects, which can be attributed to the unseen area underneath. Conversely, employing a flexible viewpoint distribution with a wider coverage angle enhances the quality of reconstruction.

\subsection{Ablation study}

\begin{table}\small
    \centering
    \begin{tabular}{@{}lcccc@{}}
   \toprule
    & PSNR$\uparrow$ & SSIM$\uparrow$ & LPIPS$\downarrow$ & Runtime$\downarrow$ \\
   \midrule
   $N=16, F=4$ & \textbf{20.49} & \textbf{0.855} & \textbf{0.128} & \textbf{12s} \\
   $N=16, F=8$ & 20.27 & 0.847 & 0.135 & 18s \\
   $N=16, F=16$ & 20.32 & 0.833 & 0.152 & 30s \\
   \midrule
   w/o light field PE & 19.83 & 0.845 & 0.139 & -- \\
   w/o ray-relative PE & 20.01 & 0.841 & 0.141 & -- \\
   \bottomrule
\end{tabular}
\caption{Ablation study under the elevation $30^{\circ}$ setting. We report PSNR, SSIM~\cite{ssim}, LPIPS~\cite{lpips} and Runtime for each ablation. Results show that our nearby views aggregation and the positional encoding lead to superior generalization performance.}
\label{tab:ablation}
\end{table}

To assess the effectiveness of our design for ECA Block (\cref{sec:eca_block}), we conducted ablation experiments on multiview synthesis at elevation $30^{\circ}$ setting (\cref{tab:ablation}). We investigated the impact of the number of neighboring views used for aggregating features in Near-Views Cross Attention module, setting $F$ to $\{4,8,16\}$, where $F=16$ indicates the use of all perspective features aggregated to the target view. Results show that increasing $F$ consumes more time and reduces the quality of the generated multiview structure and appearance. Additionally, we examined the positional encoding (PE) strategies and found that removing either the light field PE or the ray-relative PE leads to a decrease in the consistency and quality of the generated images.

\section{Limitations and Conclusion}
\label{sec:conclusion}

\cparagraph{Limitations and future works} The proposed EpiDiff still has the following limitations. Firstly, EpiDiff's efficacy in generating multiview images from arbitrary viewpoints is less pronounced compared to closer views. This is primarily due to the instability of the base NVS model when generating novel views significantly distant from the input view. We anticipate the emergence of more robust base models and enhanced stability in multiview synthesis upon integration with EpiDiff.
Secondly, the process of generating a 3D model from a single image is divided into two separate steps: multiview synthesis and 3D reconstruction. Notably, the acquisition of 3D knowledge is an integral component of both these stages.
Therefore, a unified pipeline, trained on extensive 2D image datasets and 3D datasets, would offer a marked improvement in efficiency and generalizability.

\cparagraph{Conclusion} In this paper, we present EpiDiff, a localized interactive multiview diffusion model designed for generating multiview-consistent and high-quality images from a single-view image. With the Zero123~\cite{zero123} as the backbone, EpiDiff adopts a lightweight epipolar attention block in the UNet, which utilizes epipolar contraints to enable cross-view interaction among feature maps of neighboring views. This module models 3D consistency whinin the original feature space of the diffusion model, exhitbiting adaptability to various base diffusion models. Extensive experiments demonstrate that EpiDiff not only efficiently generates consistent multiview images, but is also capable of generating more freely distributed views, help to better reconstruction from generated multiviews.

\section*{Acknowledgment}

This work was supported by National Natural Science Foundation of China (62132001).

{
    \small
    \bibliographystyle{ieeenat_fullname}
    \bibliography{main}
}


\end{document}